\documentclass[conference]{IEEEtran}
\IEEEoverridecommandlockouts
\usepackage{soul}
\usepackage{cite}
\usepackage{amsmath,amssymb,amsfonts}
\usepackage{algorithmic}
\usepackage{graphicx}
\usepackage{textcomp}
\usepackage{balance}
\usepackage{xcolor}

\usepackage{hyperref}
\usepackage{multirow}
\def\BibTeX{{\rm B\kern-.05em{\sc i\kern-.025em b}\kern-.08em
    T\kern-.1667em\lower.7ex\hbox{E}\kern-.125emX}}
\begin{document}

\title{Event-Driven Visual-Tactile Sensing\\and Learning for Robots
}

\author{\IEEEauthorblockN{Tasbolat Taunyazov\IEEEauthorrefmark{1}, Weicong Sng\IEEEauthorrefmark{1}, Hian Hian See\IEEEauthorrefmark{2}, Brian Lim\IEEEauthorrefmark{2}\IEEEauthorrefmark{3},\\Jethro Kuan\IEEEauthorrefmark{1}, Abdul Fatir Ansari\IEEEauthorrefmark{1}, Benjamin C.K. Tee\IEEEauthorrefmark{2}\IEEEauthorrefmark{3}, and Harold Soh\IEEEauthorrefmark{1}}
\IEEEauthorblockA{\IEEEauthorrefmark{1}\textit{Dept. of Computer Science, National University of Singapore}}
\IEEEauthorblockA{\IEEEauthorrefmark{2}\textit{Dept. of Materials Science and Engineering, National University of Singapore}\\
\IEEEauthorblockA{\IEEEauthorrefmark{3}\textit{Institute for Health Technology and Innovation, National University of Singapore}}
Email: ttaunyazov@u.nus.edu, dcssngw@nus.edu.sg,  mseshh@nus.edu.sg, brian.lim@nus.edu.sg,\\jethro@comp.nus.edu.sg, abdulfatir@u.nus.edu, benjamin.tee@nus.edu.sg, harold@comp.nus.edu.sg
}
}

\newcommand{\hsnote}[1]{[{\color{purple}HS: {#1}}]}
\newcommand{\wcnote}[1]{[{\color{red}WC: {#1}}]}
\newcommand{\hhnote}[1]{[{\color{blue}HH: {#1}}]}
\newcommand{\jknote}[1]{[{\color{green}JK: {#1}}]}
\newcommand{\ttnote}[1]{[{\color{magenta}TT: {#1}}]}

\maketitle

\begin{abstract}
This work contributes an event-driven visual-tactile perception system, comprising a novel biologically-inspired tactile sensor and multi-modal spike-based learning. Our neuromorphic fingertip tactile sensor, NeuTouch, scales well with the number of taxels thanks to its event-based nature. Likewise, our Visual-Tactile Spiking Neural Network (VT-SNN) enables fast perception when coupled with event sensors. We evaluate our visual-tactile system (using the NeuTouch and Prophesee event camera) on two robot tasks: container classification and rotational slip detection. On both tasks, we observe good accuracies relative to standard deep learning methods. We have made our visual-tactile datasets freely-available to encourage research on multi-modal event-driven robot perception, which we believe is a promising approach towards intelligent power-efficient robot systems. 
\end{abstract}

\section{Introduction}

Many everyday tasks require multiple sensory modalities to perform successfully. For example, consider fetching a carton of soymilk from the fridge~\cite{billard2019trends}; humans use vision to locate the carton and can infer from a simple grasp how much liquid the carton contains. 
They can then use their sense of sight and touch to lift the object without letting it slip. 
These actions (and inferences) are performed robustly using a power-efficient neural substrate---compared to    the multi-modal deep neural networks used in current artificial systems, human brains require far less energy~\cite{li2016,StrubellGM19}.

In this work, we take crucial steps towards efficient visual-tactile perception for robotic systems. We gain inspiration from biological systems, which are \emph{asynchronous} and \emph{event-driven}. In contrast to resource-hungry deep learning methods, event-driven perception forms an alternative approach that promises power-efficiency and low-latency---features that are ideal for real-time mobile robots. However, event-driven systems remain under-developed relative to standard synchronous perception methods~\cite{Pfeiffer2018,liu2019event}. 

This paper\footnote{The original paper version is published in the RSS 2020 proceedings. This version includes the complete appendix, updated results for power utilization on the Loihi, and new results for a larger container \& weight classification dataset.} makes multiple contributions that advance event-driven \emph{visual-tactile} perception. First, to enable richer tactile sensing, we contribute the 39-taxel \emph{NeuTouch fingertip sensor}. Compared to existing commercially-available tactile sensors, NeuTouch's neuromorphic design enables scaling to a larger number of taxels while retaining low latencies. 

Next, we investigate {multi-modal learning} with NeuTouch and the Prophesee event camera. Specifically, we develop a \emph{visual-tactile spiking neural network} (VT-SNN) that uses both sensory modalities for supervised-learning tasks. Different from conventional deep artificial neural network (ANN) models~\cite{LeCun2015}, SNNs process discrete spikes asynchronously and thus, are arguably better suited to the event data generated by our neuromorphic sensors. In addition, SNNs can be used on efficient low-power neuromorphic chips such as the Intel Loihi~\cite{Davies2018}. We train our VT-SNN using recent spike-based backpropagation~\cite{shrestha2018slayer} and introduce a \emph{weighted spike-count} loss to encourage early classification. %

Our experiments center on two robot tasks: object classification and (rotational) slip detection. In the former, we tasked the robot to determine the type of container being handled and amount of liquid held within. The containers were opaque with differing stiffness, and hence, both visual and tactile sensing are relevant for accurate classification. We show that relatively small differences in weight ($\approx 30$g across 20 object-weight classes) can be distinguished by our prototype sensors and spiking models. The slip detection experiment indicates rotational slip can be accurately detected within $0.08$s (visual-tactile spikes processed every $\approx 1$ms). %
In both experiments, SNNs achieved competitive (and sometimes superior) performance relative to ANNs with similar architecture. When tested on the Intel Loihi, the SNN achieved inference speeds similar to a GPU in a real-time simulation, but consumed approximately 1900 times less power.

Taking a broader perspective, event-driven perception represents an exciting opportunity to enable power-efficient intelligent robots. This work suggests that an ``end-to-end'' event-driven perception framework is promising and warrants further research. We have made the data from our two experiments (and a third dataset that expands the number of grasped items to 36 different objects using a similar protocol) freely-available to the research community\footnote{Available at \url{https://clear-nus.github.io/visuotactile/}}. To our knowledge, these represent the first publicly-available \emph{event-based} \emph{visual-tactile} datasets, and we hope their availability will spur research on event-driven robotics. To summarize, our primary contributions are:
\begin{itemize}
    \item NeuTouch, a scalable event-based tactile sensor for robot end-effectors;
    \item A Visual-Tactile Spiking Neural Network that leverages multiple event sensor modalities and trained using a new temporally-weighted spike-count loss;
    \item Systematic experiments demonstrating the effectiveness of our event-driven perception system on object classification and slip detection, with comparisons to conventional ANN methods;
    \item Visual-tactile event sensor datasets, which also includes RGB images and proprioceptive data from the robot.
\end{itemize}

\section{Background and Related Work}
\label{sec:background}
In the following, we give a brief overview of related work on visual-tactile perception for robotics, and event-driven sensing and learning. Both areas are broad and thus, we focus on the core concepts and provide links to articles that cover these topics in greater detail.

\subsection{Visual-Tactile Perception for Robots}

General recognition of the importance of multi-modal sensing for robotics has led to innovations in both sensing and perception methods. Of late, there has been a flurry of papers on combining vision and touch sensing, e.g., \cite{sinapov2014learning,gao2016deep,li2018slip,lee2019making,lin2019learning,liu2018robotic}. 
However, work on visual-tactile learning of objects dates back to (at least) 1984 when vision and tactile data was used to create a surface description of primitive objects~\cite{allen1984surface}; in this early work, tactile sensing played a supportive role for vision due to the low resolution of tactile sensors at the time.

Recent advancements in tactile technology~\cite{shan2017robotic,sferrazza2019design} have encouraged the use of tactile sensing for more complex tasks, including object exploration~\cite{liu2016visual} and classification~\cite{soh2012online,Soh2014,taunyazov2019towards}, shape completion~\cite{varley2017visual}, and slip detection~\cite{reinecke2014experimental,bekiroglu2011learning}. One popular sensor is the BioTac; similar to a human finger, it uses textured skin, allowing vibration signatures to be used for high accuracy material and object identification and slip detection~\cite{su2015force}. The BioTac has also been used in visual-tactile learning, e.g., \cite{gao2016deep} combined tactile data with RGB images to recognize objects via deep learning. Other recent works have used the Gelsight~\cite{yuan2017gelsight}---an optical-based tactile sensor---for visual-tactile slip detection~\cite{li2018slip, calandra2018more}, grasp stability, and texture recognition~\cite{luo2018vitac}. Very recent work has used unsupervised learning to generate neural representations of visual-tactile data (with proprioception) for reinforcement learning~\cite{lee2019making}.  
Compared to the prior work above, we do not leverage synchronous sensors or conventional deep learning methods. Rather, our sensor and learning method are both event-driven.

\subsection{Event-based Perception: Sensors and Learning}

Work on event-based perception has focused primarily on vision (see \cite{Gallego2018} for a comprehensive survey). This emphasis on vision can be attributed both to its applicability across many tasks, as well as the recent availability of event cameras such as the DVS and Prophesee Onboard. Unlike conventional optical sensors, event cameras capture pixel changes asynchronously. %
There are very few event-based tactile sensors; recent variants used in robotics are optical-based sensors that are constructed by attaching event-based cameras to an elastic material~\cite{naeini2019novel, kumagai2019event}. The camera captures deformations of the elastic material when it physically interacts with an object.%

Event-based sensors have been successfully used in conjunction with deep learning techniques~\cite{Gallego2018}. The binary events are first converted into real-valued tensors, which are processed downstream by deep ANNs. This approach generally yields good models (e.g., for motion segmentation~\cite{Mitrokhin2019}, optical flow estimation~\cite{zhu2018ev}, and car steering prediction~\cite{maqueda2018event}), but at high compute cost.

Neuromorphic learning, specifically Spiking Neural Networks (SNNs)~\cite{Pfeiffer2018,Tavanaei2019}, provide a competing approach for learning with event data. Similar to event-based sensors, SNNs work directly with discrete spikes and hence, possess similar characteristics, i.e., low latency, high temporal resolution and low power consumption. Historically, SNNs have been hampered by the lack of a good training procedure. Gradient-based methods such as backpropagation were not available because spikes are non-differentiable. Recent developments in effective SNN training~\cite{shrestha2018slayer,bellec2019biologically,akrout2019deep} and the nascent availability of neuromorphic hardware (e.g., IBM TrueNorth~\cite{Merolla668} and Intel Loihi~\cite{Davies2018}) have renewed interest in neuromorphic learning for various applications, including robotics. SNNs do not yet consistently outperform their deep ANN cousins on pseudo-event image datasets, and the research community is actively exploring better training methods for real event-data. 

In this work, we develop a visual-tactile SNN trained using SLAYER~\cite{shrestha2018slayer}, a recent (approximate) backpropagation method for SNNs. There has been relatively little work on multi-modal SNNs and prior work has focused on audio-video data~\cite{chevallier2005distributed,Rathi2018} (e.g., for emotion detection~\cite{mansouri2019speech}). Very recent work~\cite{Zhou2019} has investigated turn-taking in human-robot collaboration using a range of non-neuromorphic sensors (Kinect video, EEG etc.). In contrast, we develop a multi-modal SNN for visual and \emph{tactile} event data; note that the latter is an ``active'' sense with different characteristics from audio.

\section{NeuTouch: An Event-based Tactile Sensor}

\begin{figure}
\centering
\includegraphics[width=0.80\columnwidth]{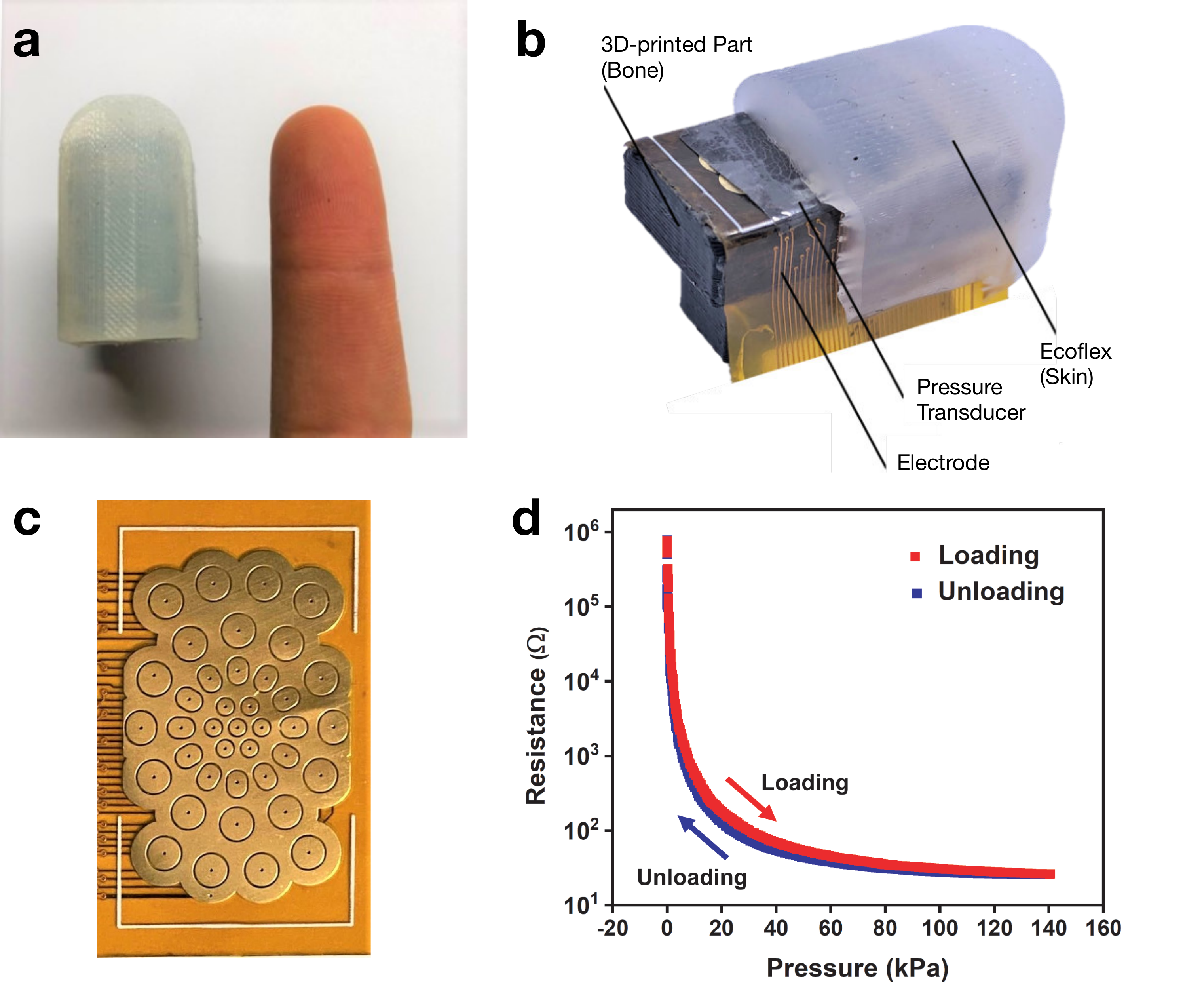}
\caption{(a) NeuTouch compared to a human finger. (b) Cross-sectional view of NeuTouch and constituent components. NeuTouch performs tactile sensing using an electrode layer with 39 taxels and a graphene-based piezoresistive thin film that is embedded beneath the protective Ecoflex ``skin''. (c) Spatial distribution of the 39 taxels on NeuTouch. (d) Pressure response curve of the graphene-based pressure transducer. }
\label{img:neutouch}
\end{figure}

Although there are numerous applications for tactile sensors (e.g., minimal invasive surgery~\cite{konstantinova2014implementation} and smart prosthetics~\cite{wu2018skin}), tactile sensing technology lags behind vision. In particular, current tactile sensors remain difficult to scale and integrate with robot platforms. The reasons are twofold: first, many tactile sensors are interfaced via time-divisional multiple access (TDMA), where individual taxels are \emph{periodically} and \emph{sequentially} sampled. The serial readout nature of TDMA inherently leads to an increase of readout latency as the number of taxels in the sensor is increased. Second, high spatial localization accuracy is typically achieved by adding more taxels in the sensor; this invariably leads to more wiring, which complicates integration of the skin onto robot end-effectors and surfaces.

Motivated by the limitations of the existing tactile sensing technology, we developed a novel Neuro-inspired Tactile sensor (NeuTouch) for use on robot end-effectors (Fig. \ref{img:neutouch}). The structure of NeuTouch is akin to a human fingertip: it comprises ``skin'', and ``bone'', and has a physical dimension of $37\times 21\times 13$ mm. This design facilitates integration with anthropomorphic end-effectors (for prosthetics or humanoid robots) and standard multi-finger grippers; in our experiments, we use NeuTouch with a Robotiq 2F-140 gripper. We focused on a fingertip design in this paper, but alternative structures can be developed to suit different applications. 

Different from earlier work~\cite{lee2019neuro}, tactile sensing is achieved via a layer of electrodes with 39 taxels and a graphene-based piezoresistive thin film. The taxels are elliptically-shaped to resemble the human fingertip’s fast-adapting (FA) mechano-receptors~\cite{johansson1978tactile}, and are radially-arranged with density varied from high to low, from the center to the periphery of the sensor. During typical grasps, NeuTouch (with its convex surface) tends to make initial contact with objects at its central region where the taxel density is the highest. Correspondingly, rich tactile data can be captured in the earlier phase of tactile sensing, which may help  accelerate inference (e.g., for early classification). The graphene-based pressure transducer forms an effective tactile sensor~\cite{sun2019fingertip,he2019universal} due to its high Young’s modulus, which helps to reduce the transducer’s hysteresis and response time. Figure~\ref{img:neutouch}d shows the pressure response of the transducer, and a low hysteresis can be observed from the loading and unloading curves.

A 3D-printed component serves the role of the fingertip bone, and Ecoflex 00-30 emulates skin for NeuTouch. The Ecoflex offers protection for the electrodes for a longer use-life and amplifies the stimuli exerted on NeuTouch. The latter enables more tactile features to be collected, since the transient phase of contact encodes much of the physical description of a grasped object, such as stiffness or surface roughness~\cite{callier2019neural}. The NeuTouch exhibits a slight delay of $\approx 300$ms when recovering from a deformation due to the soft nature of Ecoflex. Nevertheless, our experiments show this effect did not impede the NeuTouch's sensitivity to various tactile stimuli.

Compared to available tactile sensors, NeuTouch is \emph{event-based} and scales well with the number of taxels; NeuTouch can accommodate 240 taxels while maintaining an exceptionally low constant readout latency of 1ms for rapid tactile perception~\cite{lee2019neuro}. 
We achieve this by leveraging upon the Asynchronously Coded Electronic Skin (ACES) platform~\cite{lee2019neuro} --- an event-based neuro-mimetic architecture that enables \emph{asynchronous} transmission of tactile information. %
With ACES, the taxels of NeuTouch mimic the function of the FA mechano-receptors of a human fingertip, which capture dynamic pressure (i.e., dynamic skin deformations)~\cite{johansson2009coding}. FA responses are crucial for dexterous manipulation tasks that require rapid detection of object slippage, object hardness, and local curvature. 

Crucially, transmission of the stimuli information is in the form of asynchronous spikes (i.e., electrical pulses), similar to biological systems; data is transmitted by individual taxels \emph{only when necessary}. This is made possible by encoding the taxels of NeuTouch with unique electrical pulse signatures. These signatures are robust to overlap and permit multiple taxels to transmit data \emph{without specific time synchronization}. Therefore, stimuli information of all the activated taxels can be combined and propagated upstream to the decoder via \emph{a single electrical conductor}. This yields lower readout latency and simpler wiring. The decoder correlates the received pulses (i.e., the combined pulse signatures) against each taxel’s known signature to retrieve the spatio-temporal tactile information. In this work, this decoding is performed in real-time on an FPGA, rather than offline~\cite{lee2019neuro}.

\section{Visual-Tactile Spiking Neural Network (VT-SNN)}

As motivated in the introduction, the successful completion of many tasks is contingent upon using multiple sensory modalities. In this work, we focus on touch and sight, i.e., we fuse tactile and visual data from NeuTouch and an event-based camera via a spiking neural model. This Visual-Tactile Spiking Neural Network (VT-SNN) enables learning and perception using both these modalities, and can be easily extended to incorporate other event sensors. 

\begin{figure}
\centering
\includegraphics[width=0.80\columnwidth]{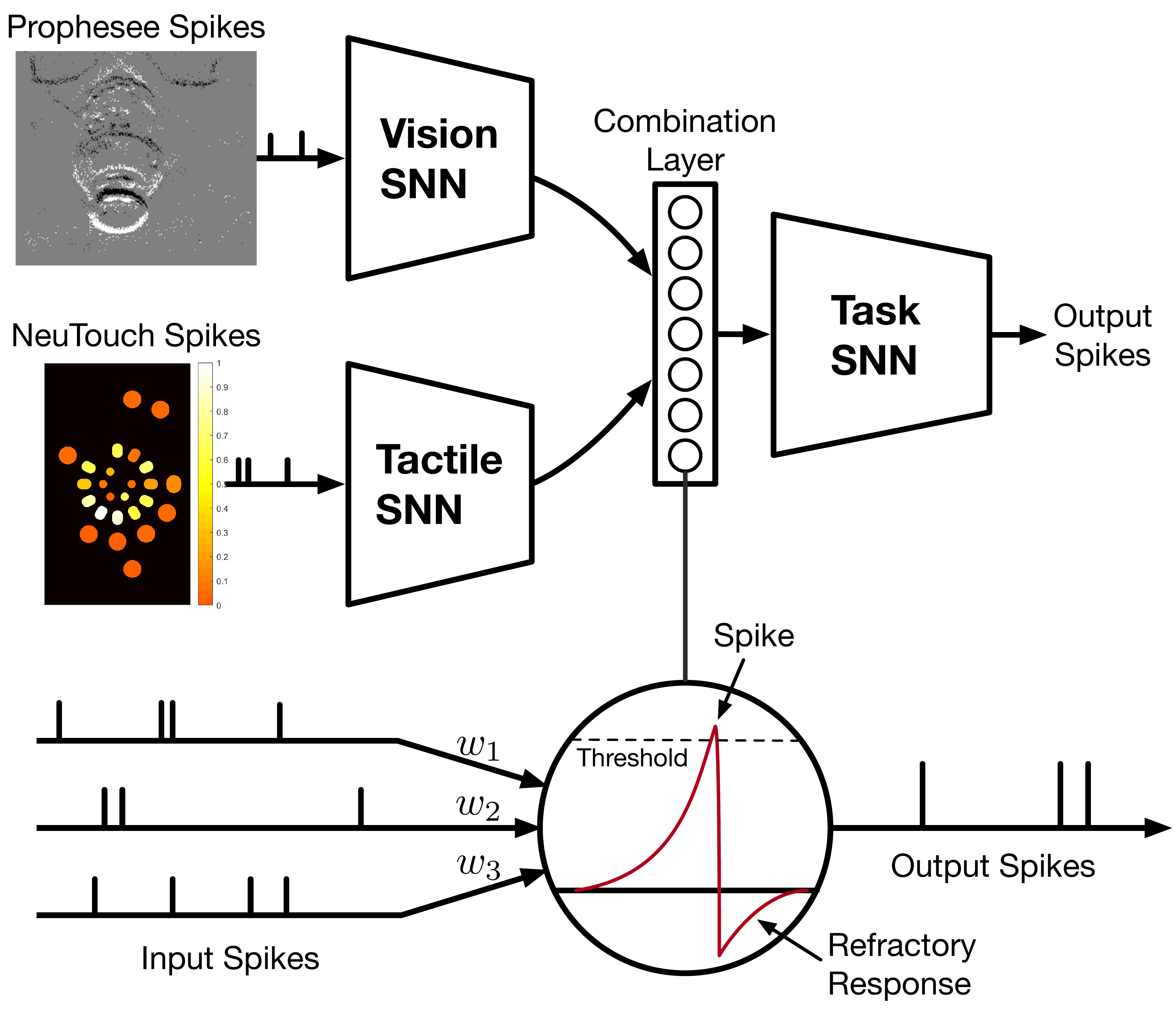}
\caption{The Visual-Tactile Spiking Neural Network (VT-SNN) which comprises two ``spiking encoders'' for each modality. The spikes from these two encoders are combined via a fixed-width combination layer and propagated to a task SNN that outputs a task-specific output spike-train. VT-SNN employs the Spike Response Model (SRM) neuron that integrates incoming spikes and spikes when a threshold is breached.}
\label{img:VT_SNN_arch_diag}
\end{figure}

\vspace{0.3em}
\noindent\textbf{Model Architecture.} From a bird's-eye perspective, the VT-SNN employs a simple architecture (Fig. \ref{img:VT_SNN_arch_diag}) that first encodes the two modalities into individual latent (spiking) representations, that are combined and further processed through additional layers to yield a task-specific output. 

We now detail the precise network structures used in our experiments, but VT-SNN may use alternative network structures for the Tactile, Vision and Task SNNs. The Tactile SNN employs a fully connected (FC) network consisting of 2 dense spiking layers\footnote{In preliminary experiments, we also tested spiking convolutional layers but it resulted in poorer performance.}. It has an input size of 156 (two fingers, each with the 39 taxels with a positive and negative polarity channel per taxel) and a hidden layer size of 32. The Vision SNN uses 3 layers; the first layer is a  pooling layer with kernel size and stride length of 4. The pooled spike train is passed as input to a 2-layer FC architecture identical to the Tactile SNN. The tactile and vision encoders have output sizes of 50 and 10, respectively\footnote{Several different dimension sizes were tested and a 50-10 encoding gave the best results.}. The encoded spike-trains of both modalities are concatenated, and passed into a dense spiking layer that generates output spikes. Note that the output dimensionality is dependent on the task: 20 for container \& weight classification, and 2 for rotational slip classification. The model architectures are agnostic to the size of the input time dimension. %

\vspace{0.3em}
\noindent\textbf{Neuron Model.}
We use the Spike Response Model (SRM)~\cite{gerstner1995time,shrestha2018slayer}. In the SRM, spikes are generated whenever a neuron's internal state (``membrane potential'') $u(t)$ exceeds a predefined threshold $\varphi$. Each neuron's internal state is affected by incoming spikes and a refractory response: 
\begin{align}
u(t) = \sum w_i(\epsilon * s_i)(t) + (\nu * o)(t) 
\end{align}
where $w_i$ is a synaptic weight, $*$ indicates convolution, $s_i(t)$ are the incoming spikes from input $i$,  $\epsilon(\cdot)$ is the response kernel, $\nu(\cdot)$ is the refractory kernel, and $o(t)$ is the neuron's output spike train. In words, incoming spikes $s_i(t)$ are convolved with a response kernel $\epsilon(\cdot)$ to yield a spike response signal that is scaled by a synaptic weight $w_i$.

\vspace{0.3em}
\noindent\textbf{Model Training.}
We optimized our spiking networks using SLAYER~\cite{shrestha2018slayer}. As mentioned in Sec. \ref{sec:background}, the derivative of a spike is undefined, which prohibits a direct application of backpropagation to SNNs. SLAYER overcomes this problem by using a stochastic spiking neuron approximation to derive an approximate gradient, and a temporal credit assignment policy to distribute errors. SLAYER trains models ``offline'' on GPU hardware. Hence, the spiking data needs to be binned into fixed-width intervals during the training process, but the resultant SNN model can be run on neuromorphic hardware. We used a straight-forward binning process where the (binary) value for each bin window $V_w$ was 1 whenever the total spike count in that window $\sum_{w} S$ exceeded a threshold value $S_{\text{min}}$:
\begin{equation} \label{eqn:bin}
  V_w = \begin{cases}
    1 & \sum_{w} S \ge S_\text{min} \\
    0 & \text{otherwise.}
  \end{cases}
\end{equation}

Similar to prior work~\cite{shrestha2018slayer}, class prediction is determined by the number of spikes in the output layer spike train; each output neuron is associated with a specific class and the neuron that generates the most spikes represents the winning class. The \emph{spike-count} loss is given by:
\begin{align}\label{eq:spikeCount}
    \mathcal{L} = \frac{1}{2} \sum_{n=0}^{N_o} \left( \sum_{t=0}^{T} \mathbf{s}^n(t) -  \sum_{t=0}^{T} \mathbf{\Tilde{s}}^n(t) \right)^2
\end{align}
which captures the difference between the observed output spike count  $\sum_{t=0}^{T} \mathbf{s}(t)$  and the desired spike count $\sum_{t=0}^{T} \mathbf{\Tilde{s}}(t)$ across the output neurons (indexed by $n$).%

We introduce a generalization of the spike-count loss above to incorporate temporal weighting:
\begin{align}\label{eq:weightedSpikeCount}
    \mathcal{L}_\omega = \frac{1}{2} \sum_{n=0}^{N_o} \left( \sum_{t=0}^{T} \omega(t) \mathbf{s}^n(t) -  \sum_{t=0}^{T} \omega(t) \mathbf{\Tilde{s}}^n(t) \right)^2.
\end{align}
We refer to $\mathcal{L}_\omega$ as the \emph{weighted spike-count} loss. In our experiments, we set  $\omega(t)$ to be monotonically decreasing, which encourages \emph{early classification} by down-weighting later spikes. Specifically, we used a simple quadratic function, $\omega(t) = \beta t^2 + \gamma$ with $\beta < 0$, but other forms may be used. 

For both $\mathcal{L}$ and $\mathcal{L}_\omega$, appropriate counts have to be specified for the correct and incorrect classes and are task-specific hyperparameters. We tuned them manually and found that setting the positive class count to $\approx 50\%$ of the maximum number of spikes (across each input within the considered time interval) worked well. %
In initial trials, we observed that training solely with the losses above led to rapid over-fitting and poor performance on a validation set. We explored several techniques to mitigate this issue (e.g., $\ell_1$ regularization and dropout), but found that simple $\ell_2$ regularization led to the best results.

\section{Robot and Sensors Setup}

\begin{figure}
\centering
\includegraphics[width=0.90\columnwidth]{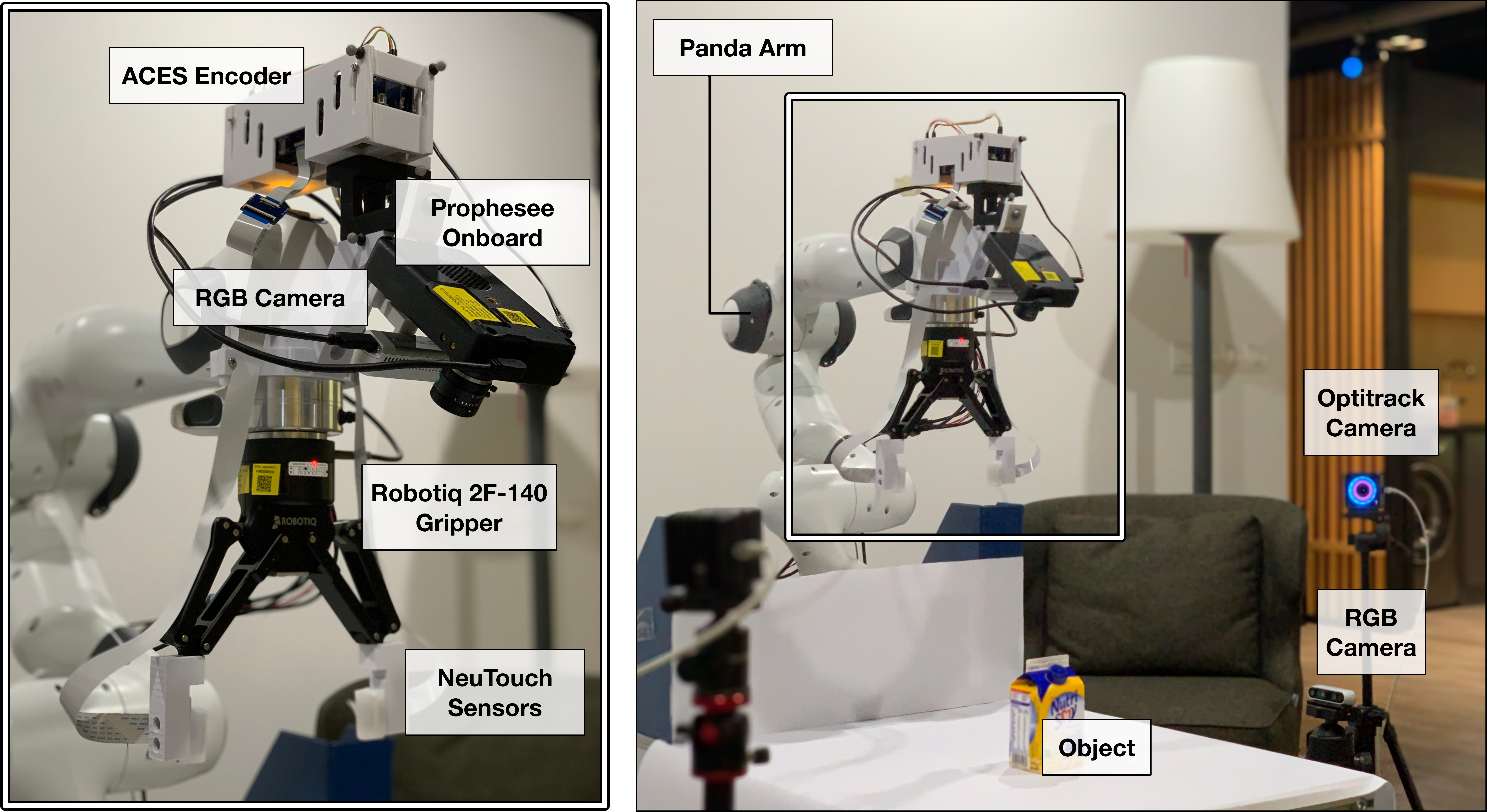}
\caption{Robot Experiment Setup. (\textbf{Left}) Close-up of the Franka Emika Panda arm and sensors; a NeuTouch sensor was attached on each Robotiq gripper finger. The Prophesee event and Realsense cameras were mounted on the arm and pointed towards the center of the gripper's grasp area. Our prototype ACES encoder was mounted on top of the arm's control panel. (\textbf{Right}) A view of the object classification experiment showing three OptiTrack cameras (11 were used, and out of scene), the RGB scene camera, and the object (soy milk carton) to be grasped and lifted.}
\label{fig:robotsetup}
\end{figure}

In this section, we describe the robot hardware setup used across our experiments (Fig. \ref{fig:robotsetup}). We used a 7-DoF Franka Emika Panda arm with a Robotiq 2F-140 gripper and collected data from four primary sensors types: NeuTouch, Prophesee Onboard, RGB cameras, and the Optitrack motion capture system. The latter two are non-event sensors and their data streams were not used in VT-SNN.

\vspace{0.3em}
\noindent\textbf{NeuTouch Tactile Sensor.}
We mounted two NeuTouch sensors to the Robotiq 2F-140 gripper and the ACES encoder on the Panda arm (Fig. \ref{fig:robotsetup}, left). To ensure consistent data, we performed a sensor warm-up before each data collection session and obtained baseline results to check for sensor drift. Specifically, we repeated 100 cycles of: closing the gripper onto a flat stiff object (the `9 hole peg test' from the YCB dataset~\cite{ycb2015}) for 3 seconds, opening the gripper, and pausing for 2 seconds. We then collected a set of benchmark data, i.e., 20 repetitions of closing the gripper onto the same `9 hole peg test' for 3 seconds. Throughout our experiments, we periodically tested for sensor drift by repeating the closing test on the `9 hole peg test' and then examining the sensor data; no significant drift was found throughout our experiments.

\vspace{0.3em}
\noindent\textbf{Prophesee Event Camera.} 
Event-based vision data was captured using the Prophesee Onboard.
Similar to the tactile sensor, each camera pixel fires asynchronously and a positive (negative) spike is obtained when there is an increase (decrease) in luminosity. The Onboard was mounted on the arm and pointed towards the gripper to obtain information about the object of interest (Fig. \ref{fig:robotsetup}). Although the camera has a maximum resolution of 640 x 480, we captured spikes from a cropped 200 x 250 rectangular window to minimize noise from irrelevant regions. The event camera bias parameters were tuned following recommended guidelines\footnote{See \url{https://support.prophesee.ai/portal/kb/articles/bias-tuning}} and we use the same parameters throughout all experiments. %
During preliminary experiments, we found the Onboard to be sensitive to high frequency ($\geq 100$Hz) luminosity changes; in other words, flickering light bulbs triggered undesirable spikes. To counter this effect, we used six Philips 12W LED White light bulbs mounted around the experiment setup to provide consistent non-flickering illumination.

\vspace{0.3em}
\noindent\textbf{RGB Cameras.}
We used two Intel RealSense D435s to provide additional non-event image data\footnote{The infrared emitters were disabled as they increased noise for the event camera and hence, no depth data was recorded.}. The first camera was mounted on the end-effector with the camera pointed towards the gripper (providing a view of the grasped object), and the second camera was placed to provide a view of the scene. The RGB images were used for visualization and validation purposes, but not as input to our models; future work may look into the integration of these standard sensors to provide even better model performance. %

\vspace{0.3em}
\noindent\textbf{OptiTrack.}
The OptiTrack motion capture system was used to collect object movement data for the slip detection experiment. We attached 6 reflective markers on the rigid parts of the end-effector and 14 markers on the object of interest. Eleven OptiTrack Prime 13 cameras were placed strategically around the experimental area to minimize tracking error (Fig. \ref{fig:robotsetup}, right); each marker was visible to most if not all cameras at any instance, which resulted in continuous and reliable tracking. We used Motive Body v1.10.0 for marker tracking and manually annotated the detected markers. 
Initial tests indicated the OptiTrack gave reliable position estimates with error $\leq 1$mm at 120Hz.  %

\vspace{0.3em}
\noindent\textbf{Further Details.}
In addition to the above sensors, we also collected proprioceptive data for the Panda arm and Robotiq gripper; these are not currently used in our models but can be included in future work. Additional information (including specific parameter settings, 3D-printed attachments, and multi-node data collection) is available in the online supplementary material\footnote{Available at \url{https://clear-nus.github.io/visuotactile/}}.

\begin{figure*}
\centering
\includegraphics[width=0.90\textwidth]{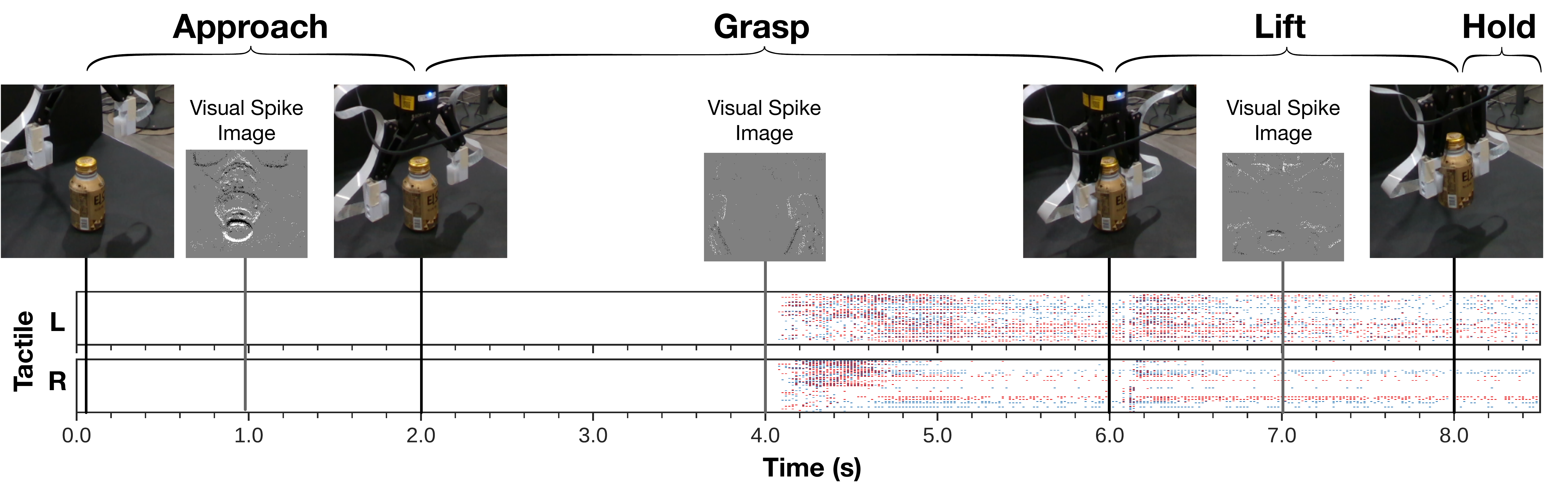}
\caption{Data collection procedure for the container \& weight classification task. The robot end-effector starts at 10 cm above the grasp point. It then \emph{approaches} the coffee can, stopping at the 2-second mark. The gripper then starts to close, \emph{grasping} it at around the 4-second mark, resulting in tactile spikes. At the 6-second mark, the robot \emph{lifts} it by 5cm above the table. The robot then holds it in the same position till the 8.5-second mark. For container and weight classification, we only use data starting from the 2-second mark.}
\label{fig:robot_motion}
\end{figure*}

\begin{figure}
\centering
\includegraphics[width=0.87\columnwidth]{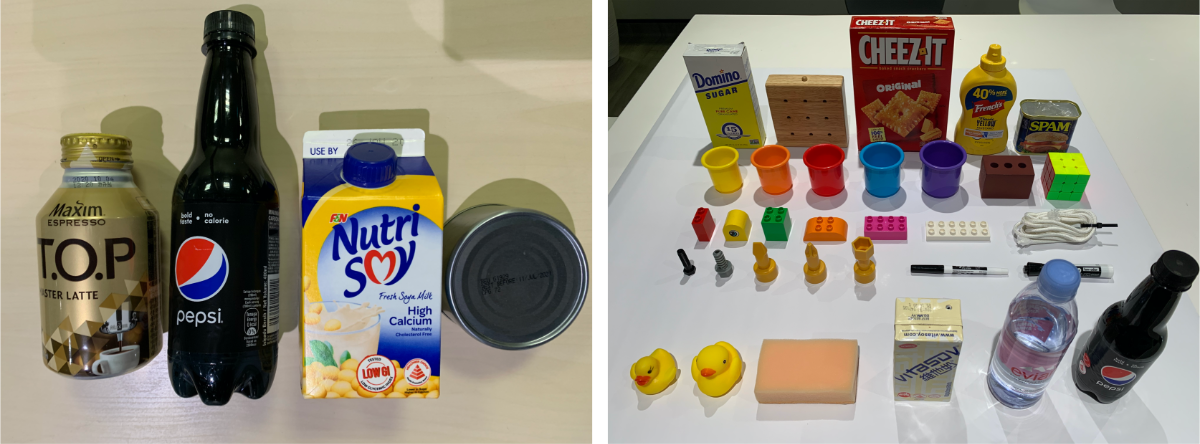}
\caption{(\textbf{Left}) Containers used for container classification task: coffee can, plastic soda bottle, soy milk carton, and metal tuna can. (\textbf{Right}) The objects in our expanded dataset ($36$ object classes with various visual and tactile profiles) collected using the same protocol.}
\label{fig:expobjects}
\end{figure}

\section{Container \& Weight Classification}
\label{sec:containerclass}

Our first experiment applies our event-driven perception framework---comprising NeuTouch, the Onboard camera, and the VT-SNN---to classify containers with varying amounts of liquid. Our primary goal was to determine (i) if our multi-modal system was effective at detecting differences in objects that were difficult to isolate using a single sensor, and (ii) whether the weight spike-count loss resulted in better early classification performance. Note that our objective was \emph{not} to derive the {best possible classifier}; indeed, we did not include proprioceptive data which would likely have improved results~\cite{lee2019making}, nor conduct an exhaustive (and computationally expensive) search for the best architecture. Rather, we sought to understand the potential benefits of using both visual and tactile spiking data in a reasonable setup. %

\subsection{Methods and Procedure}

\vspace{0.3em}
\noindent\textbf{Objects.} We used four different containers: an aluminium coffee can, a plastic Pepsi bottle, a cardboard soy milk carton and a metal tuna can (Fig. \ref{fig:expobjects}). These objects have different degrees of hardness; the soy milk container was the softest, and the tuna can was the most rigid. Because of size differences, each container was filled with differing amounts of liquid; the four objects contained a maximum of 250g, 400g, 300g, and 140g, respectively\footnote{The can did not have a cover and we filled it with a packet of rice to avoid spills and possible liquid damage. The tuna can was placed with the open side facing downwards so, the rice was not visible.}. For each object, we collected data for \{0\%, 25\%, 50\%, 75\%, 100\%\} of the respective maximum amount. This resulted in 20 object classes comprising the four containers with five different weight levels each\footnote{We have an updated dataset (Version 2.0) which contains 40 samples for each class. Please see the Appendix for results and the online website for download links.}. 

\vspace{0.3em}
\noindent\textbf{Robot Motion.} The robot would grasp and lift each object class fifteen times, yielding 15 samples per class. Trajectories for each part of the motion was computed using the MoveIt Cartesian Pose Controller~\cite{coleman2014reducing}. Briefly, the robot gripper was initialized 10cm above each object's designated grasp point. The end-effector was then moved to the grasp position (2 seconds) and the gripper was closed using the Robotiq grasp controller (4 seconds). The gripper then lifted the object by 5cm (2 seconds) and held it for 0.5 seconds. %

\vspace{0.3em}
\noindent\textbf{Data Pre-processing.} 
For both modalities, we selected data from the grasping, lifting and holding phases (corresponding to the 2.0s to 8.5s window in Figure \ref{fig:robot_motion}), and set a bin duration of 0.02s (325 bins) and a binning threshold value $S_{\text{min}} = 1$. We used stratified K-folds to create 5 splits;  each split contained 240 training and 60 test examples with equal class distribution.  %

\vspace{0.3em}
\noindent\textbf{Classification Models.} 
We compared the SNNs against conventional deep learning, specifically  Multi-layer Perceptrons (MLPs) 
with Gated Recurrent Units (GRUs)~\cite{cho2014learning} and 3D convolutional neural networks (CNN-3D)~\cite{gandarias2019active}. We trained each model using (i) the tactile data only, (ii) the visual data only, and (iii) the combined visual-tactile data. Note that the SNN model on the combined data corresponds to the VT-SNN. When training on a single modality, we use Visual or Tactile SNN as appropriate. We implemented all the models using PyTorch. The SNNs were trained with SLAYER to minimize the (weighted) spike-count loss, and the ANNs were trained to minimize the cross-entropy loss using RMSProp. All models were trained for 500 epochs. Source code implementing our models is available online at \url{https://clear-nus.github.io/visuotactile/}.

\begin{table}
\centering 
\caption{Container \& Weight Classification (Entire Input Sequence): Average Accuracy with Standard Deviation in Brackets}
\label{tbl:classacc325}
\begin{tabular}{l|ccc}
 \hline 
 \hline 
  \textbf{Model} & \textbf{Tactile} & \textbf{Vision} & \textbf{Combined} \\
  \hline
 SNN ($\mathcal{L}$) & 0.71 (0.045) & 0.73 (0.064)  & 0.81 (0.039)\\
  \hline
 SNN  ($\mathcal{L}_\omega$) & 0.71 (0.023) & 0.72 (0.065)  & 0.80 (0.048)\\
  \hline 
 ANN (MLP-GRU) & 0.50 (0.059) & 0.43 (0.054) & 0.44 (0.062)\\
   \hline 
 ANN (CNN-3D) & 0.75 (0.061) & 0.68 (0.022) & 0.80 (0.041) \\
 \hline 
 \hline 
\end{tabular}
\end{table}

\subsection{Results and Analysis}

\vspace{0.3em}
\noindent\textbf{Model Comparisons.} 
The test accuracies of the models (using the entire input sequence) are summarized in Table \ref{tbl:classacc325}. 
The multi-modal SNN model achieves the highest score of 81\%, an improvement of $\approx$10\% compared to the single-modality variants. 
Using either modality alone results in comparable performance.  
We were initially surprised by the accuracy attained by the vision-only model; we expected performance $\approx 20\%$ since since the containers were easily distinguishable by sight, but the weight category was not. However, a closer examination of the data showed that (i) the Pepsi bottle was not fully opaque and the water level was observable by Onboard on some trials, and (ii) the Onboard was able to see object deformations as the gripper closed, which revealed the ``fullness'' of the softer containers. 

\begin{figure}
\centering
\includegraphics[width=0.70\columnwidth]{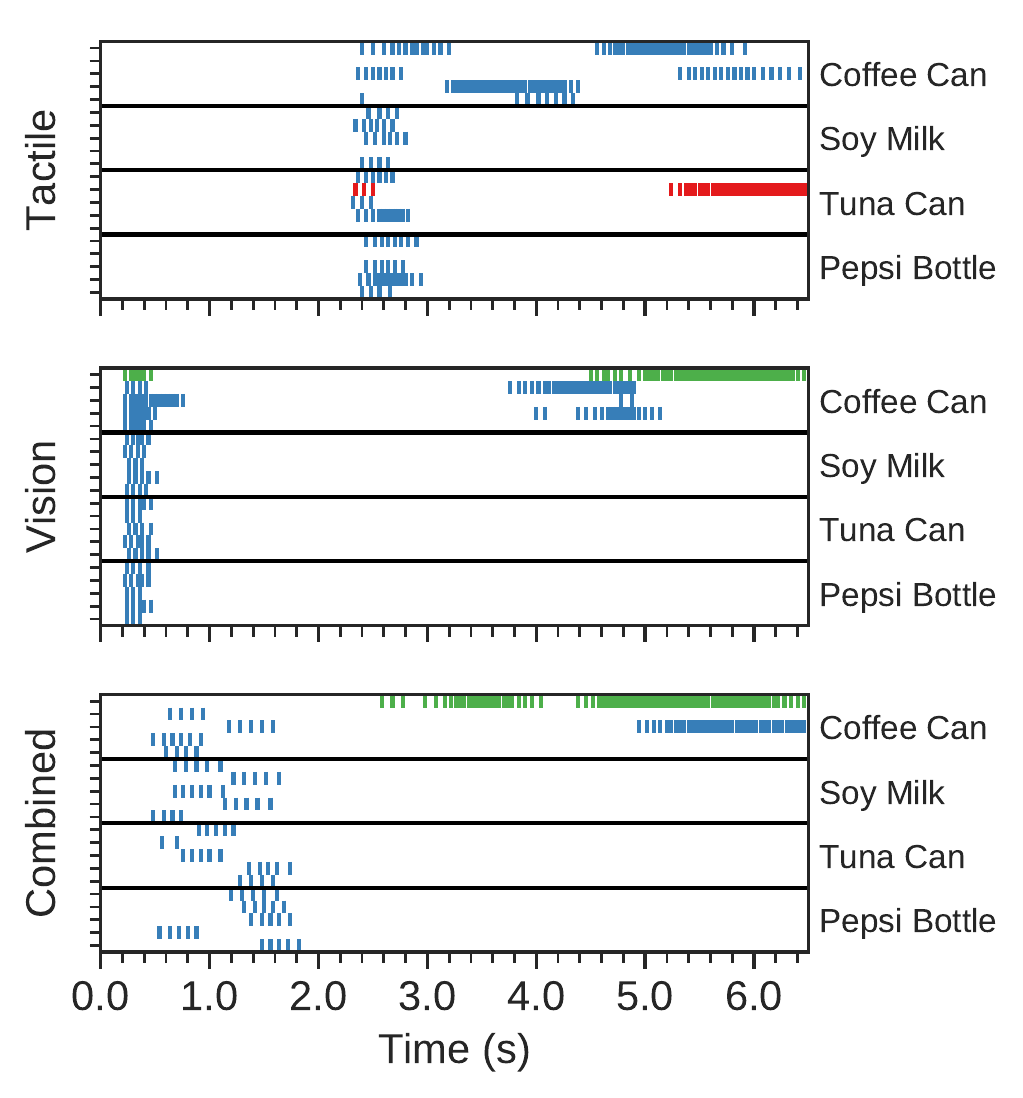}
\caption{Output spikes (blue lines) for the models trained with different modalities with correct and incorrect predictions in green and red, respectively, while grasping a coffee can with 100\% weight. The weight categories are arranged from 0\% to 100\% (bottom to top) for each container. The tactile model is unable to distinguish between a coffee can and a tuna can while the vision model is uncertain about the weight. The combined visual-tactile model predicts the correct class with high certainty.}
\label{fig:spikeclass}
\end{figure}

Figure \ref{fig:spikeclass} shows an instructive example showing the advantage of fusing both modalities. It shows the output spikes from the different SNN models for a coffee can with 100\% weight. The models trained on tactile and vision data are uncertain of the container and the weight category, respectively. We see the tactile model is unable to discern between tuna can and coffee can. On the other hand, the vision model correctly predicts the container but is unsure about the weight category. The combined visual-tactile model incorporates information from both the modalities and is able to predict the correct class (both container and weight categories) with high certainty.

The SNN models performed far better than the MLP-GRU model, and similar to the 3D CNNs. We had expected comparable performance since MLP-GRU models are known to perform well on a variety of tasks. The poor performance was possibly due to the relatively long sample durations (325 time-steps) and the large number of parameters in the ANN models, relative to the size of our dataset. %

\begin{figure}
\centering
\includegraphics[width=0.80\columnwidth]{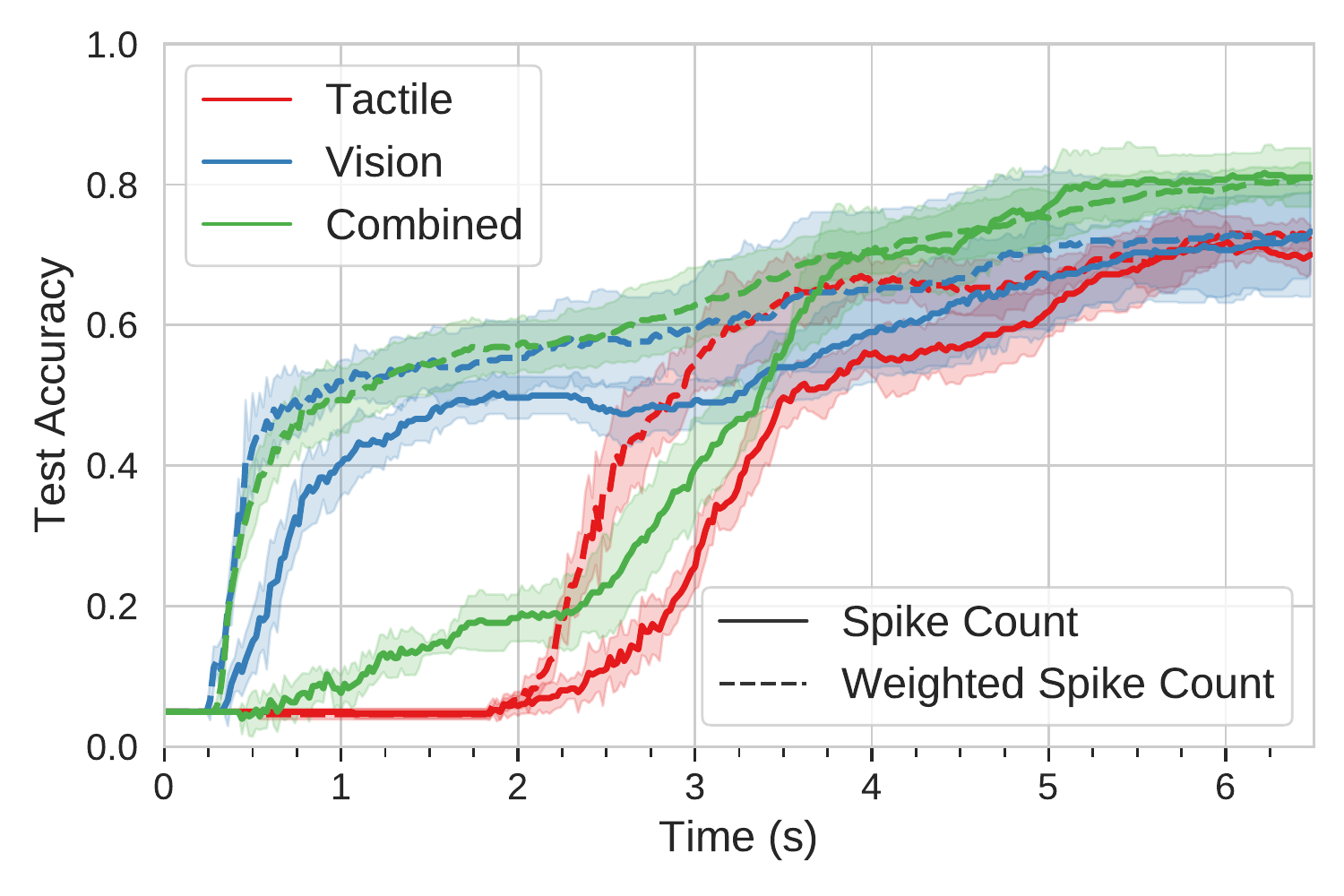}
\caption{Container and weight classification accuracy over time. Lines show average test accuracy and shaded regions represent the standard deviations. Vision-only classification results in higher early accuracy as visual spikes are obtained as the gripper is closing, and tactile events arise only upon contact with the object. Combining both vision and tactile event data results in significantly higher accuracy, compared to using each modality separately. Using the weighted spike count loss results in better early classification.}

\label{fig:classtime}
\end{figure}

\vspace{0.3em}
\noindent\textbf{Early Classification.} Instead of waiting for all the output spikes to accumulate, we can perform early classification based on the number of spikes seen up to time $t$. Fig. \ref{fig:classtime} shows the accuracies of the different models over time. Between $0.5-3.0$s, the vision model was already able to distinguish between certain objects. We posit this was due to small movements (of the mounted camera) as the gripper closed, which resulted in changes perceived by the Onboard. As expected, output tactile spikes do not emerge until after contact is made with the object at $\approx 2$s.

Although the two losses $\mathcal{L}$ and $\mathcal{L}_\omega$ have similar ``final'' accuracies in Table \ref{tbl:classacc325}, we see that $\mathcal{L}_\omega$ has a significant impact on test accuracies over time. This effect is most clearly seen for the combined visual-tactile model; the $\mathcal{L}_\omega$ variant has a similar early  accuracy profile as vision, but achieves better performance as tactile information is accumulated.

\section{Rotational Slip Classification}
\label{sec:slip}

\begin{figure}
\centering
\includegraphics[width=0.70\columnwidth]{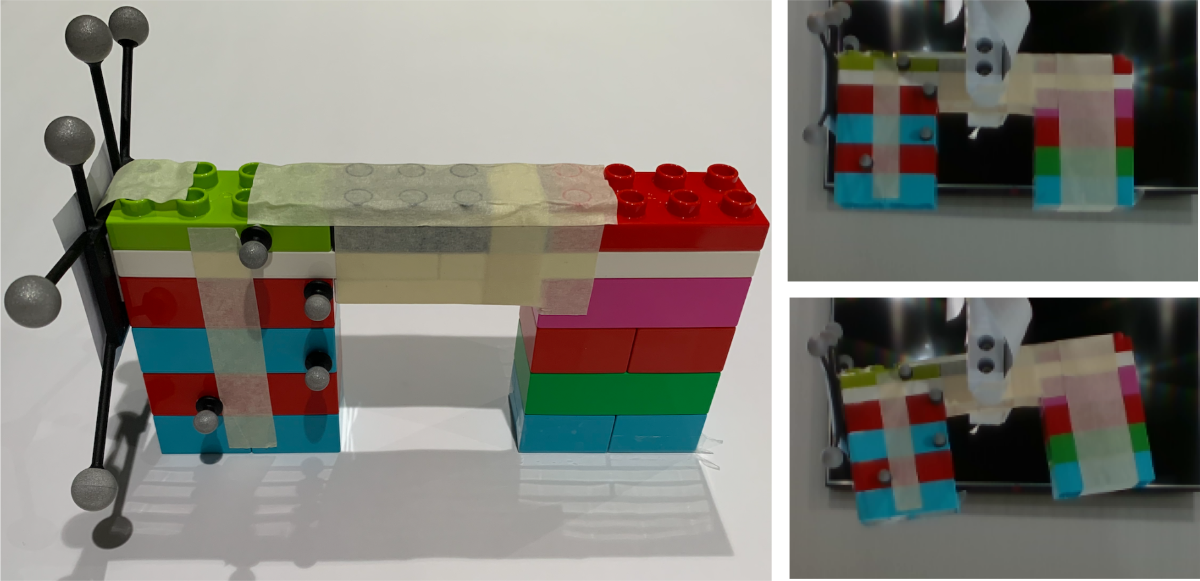}
\caption{(\textbf{Left}) Object for the Slip Classification Task with attached OptiTrack markers. (\textbf{Right}) Object during a stable grasp (top) and unstable grasp with rotational slip (bottom); the bottom object has additional mass transferred to the left leg (from the right leg) that causes it to rotate during lifting. Note that both objects had equal mass. %
}
\label{img:slips}
\end{figure}

In this second experiment, we task our perception system to classify rotational slip, which is important for stable grasping; stable grasp points can be incorrectly predicted for objects with center-of-mass that are not easily determined by sight, e.g., a hammer and other irregularly-shaped items. Accurate detection of rotational slip will allow the controller to re-grasp the object and remedy poor initial grasp locations. However, to be effective, slip detection needs to be performed accurately and rapidly.

\subsection{Method and Procedure}

\vspace{0.3em}
\noindent\textbf{Objects.} Our test object was constructed using Lego Duplo blocks (Fig. \ref{img:slips}) with a hidden 10g mass in each leg. The ``control'' object was designed to be balanced at the grasp point. To induce rotational slip, we modified the object by transferring the hidden mass from the right leg to the left. As such, the stable and unstable objects were visually identical and had the same overall weight.

\vspace{0.3em}
\noindent\textbf{Robot Motion.} The robot would grasp and lift both object variants 50 times, yielding 50 samples per class. Similar to the previous experiment, motion trajectories were computed using  MoveIt~\cite{coleman2014reducing}. The robot was instructed to close upon the object, lift by 10cm off the table (in 0.75 seconds) and hold it for an additional 4.25 seconds. We tuned the gripper's grasping force to enable the object to be lifted, yet allow for rotational slip for the off-center object (Fig. \ref{img:slips}, right).

\vspace{0.3em}
\noindent\textbf{Data Preprocessing.} Instead of training our models across the entire movement period, we extracted a short time period in the lifting stage. The exact start time was obtained by analyzing the OptiTrack data; specifically, we obtained the baseline orientation distribution (for 1 second) and defined rotational slip as a deviation larger than 98\% of the baseline frames. We found that slip occurred almost immediately during the lifting. Since we were interested in rapid detection, we extracted a 0.15s window around the start of the lift, and set a bin duration of 0.001s (150 bins) with binning threshold $S_\text{min} = 1$. We used stratified K-folds to obtain 5 splits, where each split contained 80 training samples and 20 testing samples.

\begin{table}
\centering 
\caption{Slip Detection Accuracy (Entire Input Sequence): Average Accuracy with Standard Deviation in Brackets}
\label{tbl:sdacc}
\begin{tabular}{l|ccc}
 \hline 
 \hline 
  \textbf{Model} & \textbf{Tactile} & \textbf{Vision} & \textbf{Combined} \\
  \hline
 SNN ($\mathcal{L}$) & 0.82 (0.045) & 1.00 (0.000)  & 1.00 (0.000)\\
  \hline
 SNN  ($\mathcal{L}_\omega$) & 0.91 (0.020) & 1.00 (0.000)  & 1.00 (0.000)\\
  \hline 
 ANN (MLP-GRU) & 0.87 (0.059) & 1.00 (0.000) & 1.00 (0.000)\\
  \hline 
 ANN (CNN-3D) & 0.44 (0.086) & 0.55 (0.100) & 0.77 (0.117) \\
 \hline 
 \hline 
\end{tabular}
\end{table}

\vspace{0.3em}
\noindent\textbf{Classification Models.} The model setup and optimization procedure were identical to the previous task, with slight modifications. The output size was reduced to 2 given the binary labels and the sequence length for the MLP-GRUs was set to 150. Finally, the desired true and false spike counts were set to 80 and 5, respectively. %

\subsection{Results and Analysis}

\vspace{0.3em}
\noindent\textbf{Model Comparisons.}
Test accuracy scores are shown in Table \ref{tbl:sdacc}.
For both the SNN and MLP-GRU, the vision and multi-modal models achieve perfect accuracy using the entire input sequence; this was unsurprising as rotational slip produced a visually distinctive signature. Using only tactile events, the SNN and MLP-GRU achieved 91\% (with $\mathcal{L}_\omega$) and 87\% accuracy, respectively. The CNN-3D performed poorly for this task, possibly due to overfitting. %

\begin{figure}
\centering
\includegraphics[width=0.80\columnwidth]{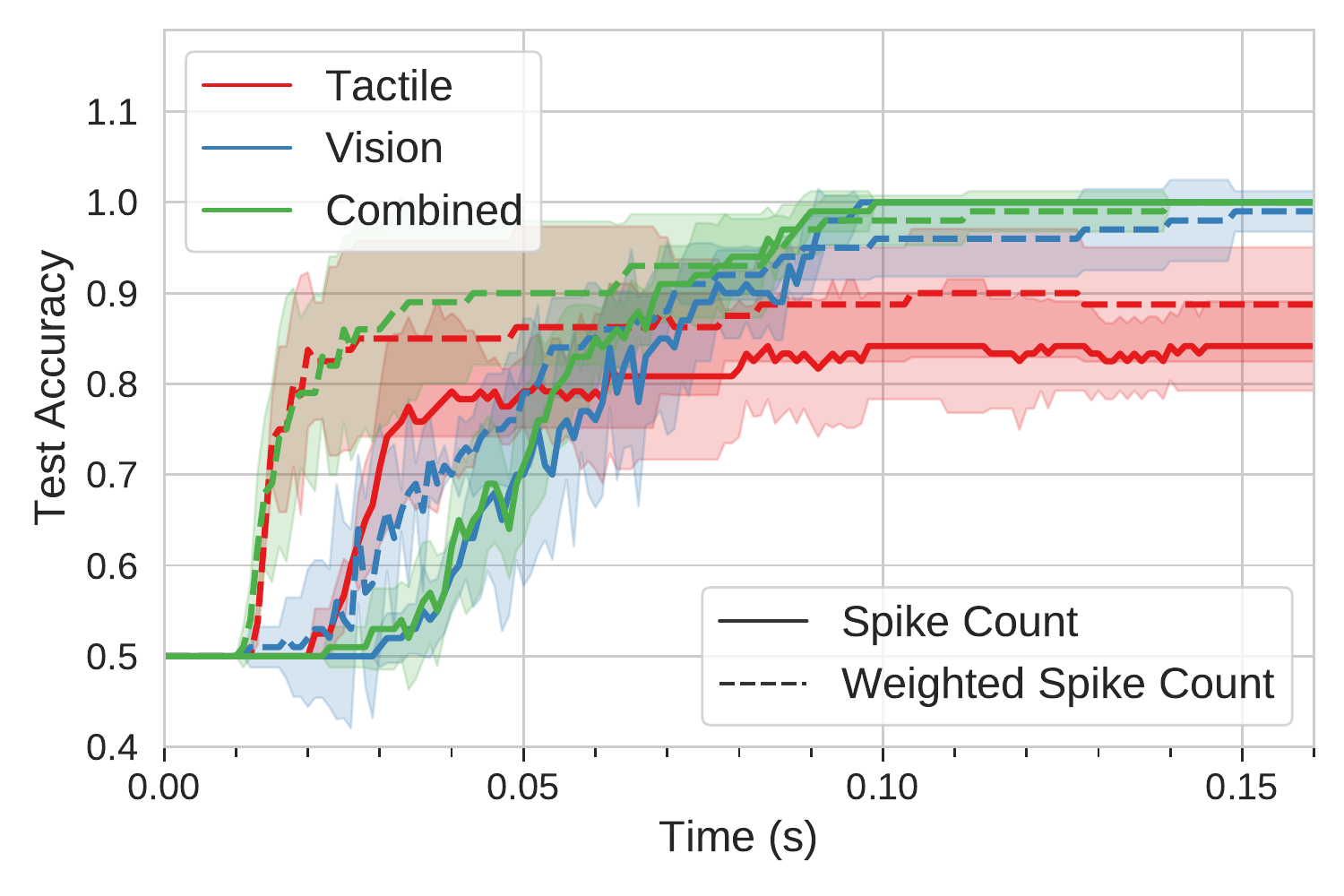}
\caption{Slip classification accuracy over time. Lines show average test accuracy and shaded regions represent the standard deviations. Classification using tactile only data results in higher early accuracy, but vision-only classification becomes more accurate as sensory data is accumulated. Combining both modalities results in higher accuracy, and the models trained with  weighted spike count loss achieves better early classification.}
\label{fig:sliptime}
\end{figure}

\vspace{0.3em}
\noindent\textbf{Early Slip Detection.} Fig. \ref{fig:sliptime} summarizes slip test accuracy at different time points. The object starts being lifted at approximately 0.01s, and we see that by 0.1s, the multi-modal VT-SNN is able to classify slip perfectly. Again, we see vision and touch possess different accuracy profiles; tactile-only classification is more accurate early on (between $0.01-0.05$s), while vision-based classification is better after $\approx 0.6$s. Fusing both modalities (under $\mathcal{L}$) results in an accuracy profile similar to vision but shifted towards higher accuracies. As before, early classification performance is significantly improved when using the weighted spike-count loss $\mathcal{L}_w$.

\section{Discussion: Speed and Power Efficiency}

We ran two benchmarking experiments comparing the VT-SNN on a Nvidia RTX 2080Ti and the neuromorphic Intel Loihi chip: (1) offline batch processing and (2) a simulated real-world setting. In the interest of brevity, we discuss the experimental setup and summarize the results for the real-world simulation setting here, and detail the procedure and full benchmark results in the appendix.

The models were tasked to perform forward passes on 1000 samples. Each sample has a length of 0.15s, which were binned every 1ms forming 150 time steps. To simulate the real-world setting, data is presented to the hardware in the different ways. For the Loihi, the model was artificially slowed down by the primary x86 core to match the 1ms timestep of the slip data. For the GPU, an artificial delay of 0.15s is introduced during the dataset fetch, to simulate the GPU waiting the length of the full window before being able to perform the inference.

The benchmark results are shown in Table \ref{tbl:benchmark-summary}, where latency is the time taken to process 1 timestep. We observe that the latency on the Loihi is slightly lower, because it is able to perform the inference as the spiking data arrives. The power consumption on the Loihi is significantly (1900x) lower than on the GPU.

\begin{table}
\centering 
\caption{Latency and Power Utilization (Real-world Simulation)}
\label{tbl:benchmark-summary}
\begin{tabular}{l|cc}
 \hline 
 \hline 
  \textbf{Hardware} & \textbf{Latency ($\mu s$)} & \textbf{Total Power ($mW$)} \\
  \hline
 Loihi & 1039.9 & 32.3 \\
  \hline
 GPU & 1045.6 & 61930 \\
 \hline 
 \hline 
\end{tabular}
\end{table}

\section{Conclusion}

In this work, we propose an event-based perception framework that combines vision and touch to achieve better performance on two robot tasks. In contrast to conventional synchronous systems, our event-driven framework asynchronously processes discrete events.
Our results suggest that the event-driven paradigm is a promising line of enquiry; we believe event-based sensing and learning will form essential parts of next-generation real-time autonomous robots that are power-efficient. We hope that the results in this paper and our datasets will encourage research in this area.

\section*{Acknowledgements}
This work was supported by the SERC, A*STAR, Singapore, through the National Robotics Program under Grant No. 172 25 00063, a NUS Startup Grant 2017-01, and the Singapore National Research Foundation (NRF) Fellowship NRFF2017-08. Thank you to Intel for access to the Neuromorphic Research Cloud, and Garrick Orchard in particular for providing valuable guidance in performing the latency and power utilization benchmarks.


\clearpage
\nobalance
\section*{Appendix}

\subsection{Event camera biases}
We tuned the hardware bias settings of Prophesee Onboard in order to minimize noise and latency, %
while retaining a good amount of signal. Table \ref{tbl:biases} shows selected key biases using Prophesee's conventions; note that the parameter values are unitless. A  file containing the full set of biases is available online at \href{https://clear-nus.github.io/visuotactile/}{https://clear-nus.github.io/visuotactile/}.

\begin{table}[h]
\centering 
\caption{Prophesee Biases}
\label{tbl:biases}
\begin{tabular}{ccc}
 \hline 
 \hline 
   \textbf{Bias} & \textbf{Value} & \textbf{Remarks}  \\
  \hline
 bias\_fo & 1775 & Pixel low-pass cut-off frequency \\ 
 \hline
 bias\_hpf & 1800 & Pixel high-pass cut-off frequency \\
 \hline
 bias\_pr & 1550 & Controls photo-receptor \\
 \hline
 bias\_diff\_on & 435 & Sensitivity to positive change in luminosity \\
 \hline
 bias\_diff\_off & 198 & Sensitivity to negative change in luminosity \\
 \hline
 bias\_refr & 1500 & Pixel refractory period  \\
 \hline
 \hline 
\end{tabular}
\end{table}

\subsection{Handling Phase Shift}
Minimizing phase shift is critical, so that machine learning models can learn meaningful interactions between the different modalities. Our setup spanned across multiple machines, each having an individual Real Time Clock (RTC). We used \textit{chronyd} to sync our clocks to the Google Public NTP pool time servers. During data collection, for each machine, we logged the record-start time according to its own RTC, and thus were able to retrieve differences between the different RTCs and sync them accordingly during data pre-processing.

\subsection{Ground-truth Slip detection}

\begin{figure} 
\centering
\includegraphics[width=1\columnwidth]{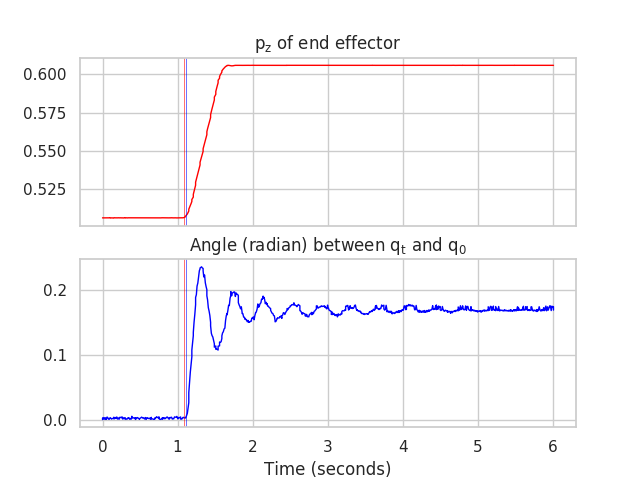}
\caption{(\textbf{Top}) $p_z$ of end-effector across time. As the robot arm lifts the object up, $p_z$ increases. (\textbf{Bottom}) $\theta_t$ (shortest angle in radians) computed between $q_t$ and $q_0$. This increases as the object slips. (\textbf{All}) The red vertical line indicates the point where $p_z$ increases significantly from at rest, and the blue vertical line indicates the point where $\theta_t$ increases significantly from at rest. The difference is 0.03 seconds for this data point.}
\label{fig:optitrack}
\end{figure}

In our data collection procedure, rotational slip typically happened in the middle of a recording. In order to extract the relevant portion of the data when slip occurred, we first detected and annotated the slip onset. We attached OptiTrack markers on Panda's end-effector and the object, such that the OptiTrack was able to determine their poses. Fig. \ref{fig:optitrack} visualizes the OptiTrack data for a typical slipping data point. We annotated the OptiTrack frame $f_{lift}$ when the robot first lifted the object up using the following heuristic:
\[   
\frac{1}{120}\sum_{i=1}^{120} I(p_{z,j}>p_{z,i}) > 0.98 = 
     \begin{cases}
       \text{False} &\quad\text{}j = f_{lift} - 1\\
       \text{True} &\quad\text{}j \geq f_{lift} \\ 
     \end{cases}
\]
We checked when $p_z$ departed the empirical noise distribution within $f_{1,\dots,120}$ when the robot arm was stationary.

For object orientation, we computed the change in angle from at rest using $\theta_t \;=\; \cos^{-1}\bigl(2\langle q_0,q_t\rangle^2 -1\bigr)$, where $q_0$ is the quaternion orientation at rest. Similarly, we annotated the frame $f_{slip}$ when the object first rotates using the following heuristic:
\[   
\frac{1}{120}\sum_{i=1}^{120} I(\theta_j>\theta_i) > 0.98 = 
     \begin{cases}
       \text{False} &\quad\text{}j = f_{slip} - 1\\
       \text{True} &\quad\text{}j \geq f_{slip} \\ 
     \end{cases}
\]
We find that the time it takes for the object to rotate upon lifting was on average 0.03 seconds across all of the slipping data points. %

\subsection{3D-Printed Parts}
\begin{figure}
\centering
\begin{minipage}{0.90\columnwidth}
\centering
\includegraphics[width=0.80\columnwidth]{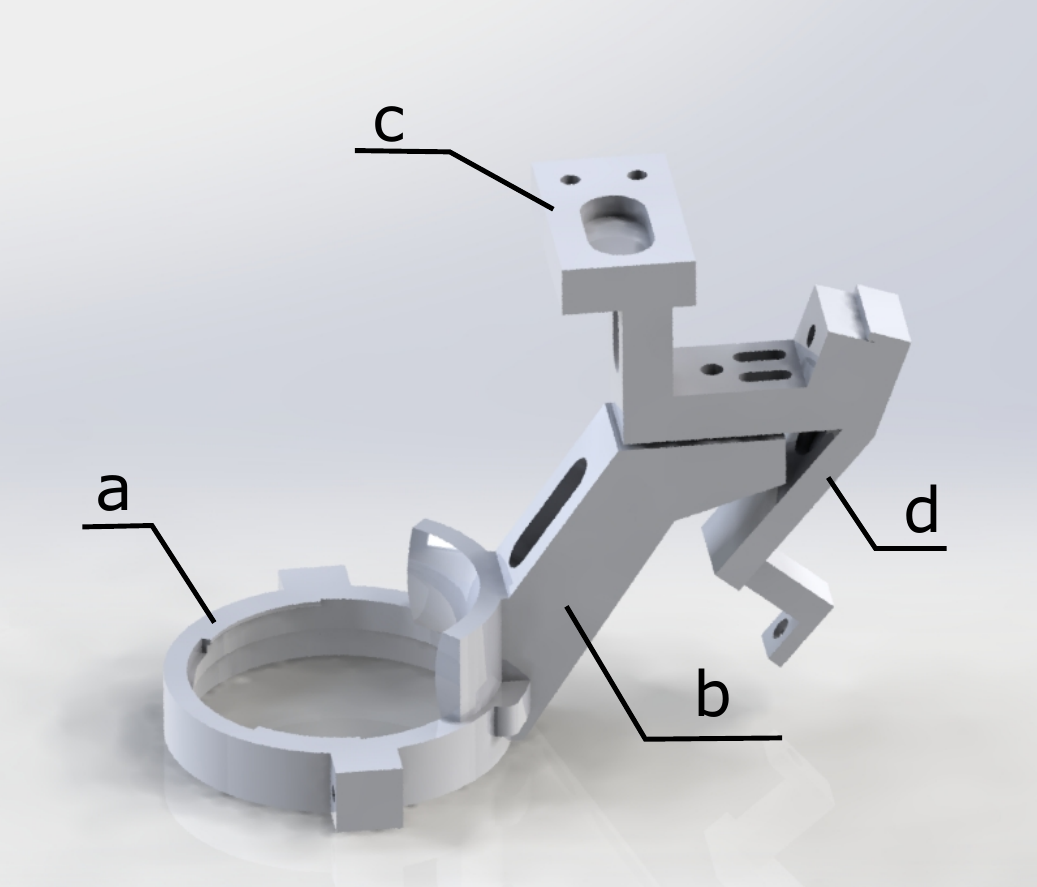}
\caption{3D-printed main holder. This 3D assembly consists of 4 parts: (a) a semi-arc to secure main holder to the 7th link of the Panda arm (infill 99\%); (b) connector to attach sensors to the Panda (infill 99\%) (c) a base for mounting the enclosure of ACES encoder (infill 80\%); (d) a holder for the Intel RealSense D435 and Prophesee Onboard (infill 80\%).}
\label{fig:main_holder}
\end{minipage}\\
\vspace{0.5cm}
\begin{minipage}{0.95\columnwidth}
\centering
\includegraphics[width=0.80\columnwidth]{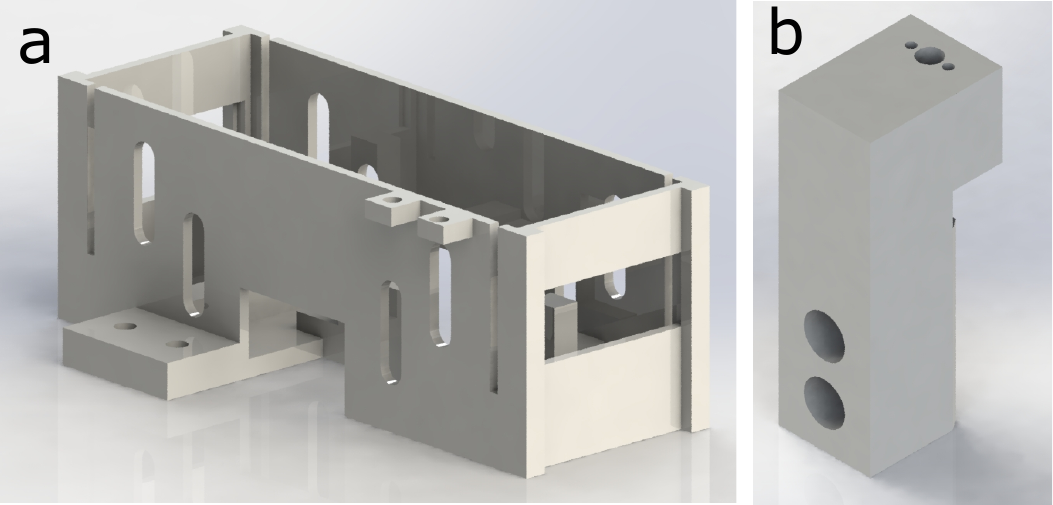}
\caption{ (a) An enclosure for the ACES encoder (infill 65\%); (b) A coupler for NeuTouch (infill 99\%). }
\label{fig:second_holder}
\end{minipage}
\end{figure}

We mounted the visual-tactile sensor components to the robot via 3D printed parts (Fig.\ref{fig:robotsetup}). There are three main 3D printed parts; a main holder (Fig.\ref{fig:main_holder}) to mount Intel RealSense D435, Prophesee Onboard and ACES encoder to the Franka Emika Panda arm, an enclosure for the ACES encoder (Fig.\ref{fig:second_holder}-a) and a coupler to mount the NeuTouch fingers onto Robotiq 2F-140 (Fig.\ref{fig:second_holder}-b).
All of the 3D printed parts for our project were printed using Acrylonitrile Butadiene Styrene (ABS) with layer thickness set to $0.2$ mm. We minimized total weight while maintaining structural integrity by maximizing the infills of only a select few components. All 3D components are available online at \href{https://clear-nus.github.io/visuotactile/}{https://clear-nus.github.io/visuotactile/}.

\subsection{Container \& Weight Dataset Version 2.0}
Since the publication of this paper, we have collected a second larger version of the weight container dataset with 40 samples per class (Version 2). For this dataset, we used a slightly different version of the  NeuTouch, which was manufactured to be more physically robust. The objects order during data collection was also randomized to mitigate any possible drift.

\begin{figure}
\centering
\includegraphics[width=0.80\columnwidth]{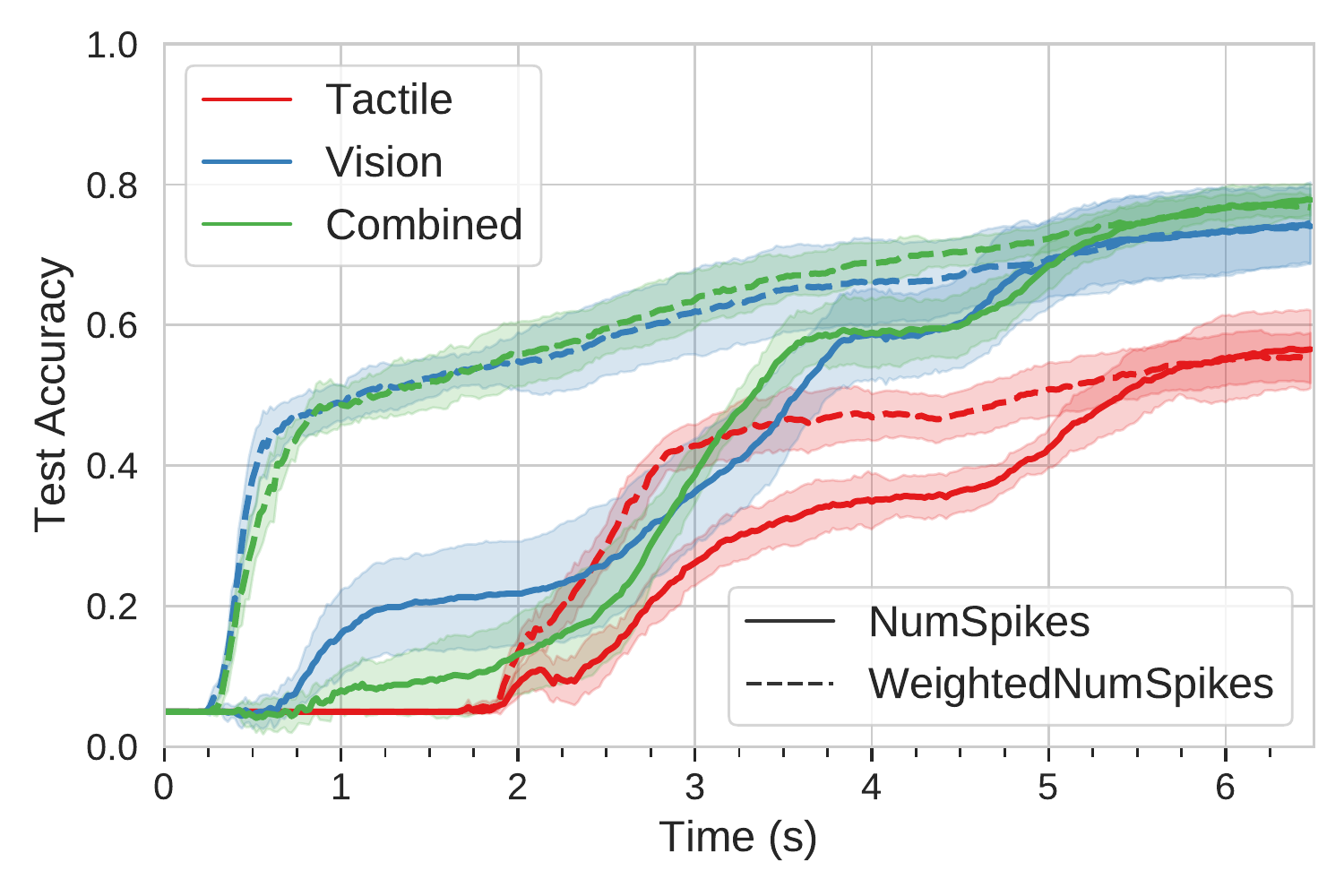}
\caption{Container and weight classification accuracy over time for Version 2 of the dataset.}

\label{fig:classtime_aug13data}
\end{figure}

We ran the VT-SNN on this dataset, with binning thresholds $S_{\min}=0$ for tactile, and $S_{\min}=1$ for vision. The obtained final accuracies are given in Table~\ref{tbl:classacc325_aug13data}. We also show early classification of the our models across time in Fig~\ref{fig:classtime_aug13data}. Overall, the findings are qualitatively similar to the previous dataset, i.e., the combined model still performs the best (albeit with a smaller improvement over vision). The weighted spike-count loss improves the overall performance of the models in the early phases of the classification. 

\begin{table}
\centering 
\caption{Container \& Weight Classification (Entire Input Sequence): Average Accuracy with Standard Deviation in Brackets}
\label{tbl:classacc325_aug13data}
\begin{tabular}{l|ccc}
 \hline 
 \hline 
  \textbf{Model} & \textbf{Tactile} & \textbf{Vision} & \textbf{Combined} \\
  \hline
 SNN ($\mathcal{L}$) & 0.57 (0.055) & 0.74 (0.054)  & 0.78 (0.022)\\
  \hline
 SNN  ($\mathcal{L}_\omega$) & 0.55 (0.036) & 0.74 (0.058)  & 0.77 (0.018)\\
  \hline 
 ANN (MLP-GRU) & 0.46 (0.048) & 0.45 (0.048) & 0.47 (0.030)\\
   \hline 
 ANN (CNN-3D) & 0.67 (0.050) & 0.65 (0.015) & 0.69 (0.042) \\
 \hline 
 \hline 
\end{tabular}
\end{table}

\subsection{Power Utilization and Latency Benchmarks}

We trained the multi-modal VT-SNN using the SLAYER framework for the task of rotational slip detection\footnote{Our benchmarking scripts for the GPU are made available in our code repository.}. The model and experimental setup is identical to rotational slip detection described in Section ~\ref{sec:slip}, with two changes:

\begin{enumerate}
    \item The Loihi neuron model is used in place of the SRM neuron model.
    \item The polarity of the vision output is discarded to reduce the vision input size to fit into a single core on the Loihi.
\end{enumerate}

Both models attain 100\% test accuracy, and produce identical results on the Loihi and the GPU. All benchmarks were obtained for the Loihi using NxSDK version 0.9.5 on a Nahuku 32 board, and on a Nvidia RTX 2080Ti GPU respectively.

\vspace{0.3em}
\noindent\textbf{Task.}
The model is tasked to perform 1000 forward passes, with a batch size of 1 on the GPU. The dataset of 1000 samples is obtained by repeating samples from our test set. Each sample consists of 0.15s of spike data, binned every 1ms into a 150 timesteps.

\vspace{0.3em}
\noindent\textbf{Latency.}
On the GPU, the system clock on the CPU was used to capture the start ($t_{start}$) and end time ($t_{end}$) for model inference, and on the Loihi, we used the system clock on superhost. We compute the latency per timestep as $(t_{end} - t_{start}) / (1000 \times 150)$, dividing across 1000 samples, each with 150 timesteps.

\vspace{0.3em}
\noindent\textbf{Power Utilization.}
To obtain power utilization on the GPU, we adopt the approach in~\cite{blouw2018benchmarking} and used the NVIDIA System Management Interface, logging \texttt{(timestamp, power\_draw)} pairs at 200ms intervals with the utility. We extracted the power draw during the time spent, and averaged it to obtain the average power draw under load. To obtain the idle power draw of the GPU, we logged power usage on the GPU for 15 minutes with no processes running on the GPU, and averaged the power draw over the period.

We use the performance profiling tools available within NxSDK 0.9.5 to obtain the power utilization for the VT-SNN on the Loihi. Our model is small and occupies less than 1 chip on the 32-chip Nahuku 32 board. To obtain more accurate power measurements, we replicate the workload 32 times and report the results per-copy. The replicated workload occupies 594 neuromorphic cores and 5 x86 cores, with 624 neuromorphic cores powered for barrier synchronization.

For the first experiment, the data was passed directly to the models. The benchmark results are presented in Table~\ref{tbl:benchmark-exp1}. We observe that the model performed inference about 10x slower on the Loihi as compared to the GPU. This was due to the data being presented faster than real time. Here, the Loihi was bottlenecked by the speed of spike injection into the x86 cores. In a more practical setting, spike data would arrive at intervals of 1ms, and the Loihi would be able to process them as they arrive, while the GPU would have to accumulate data for the full 0.15s window before performing the forward pass.

\begin{table*}
\centering
\caption{Inference Speed and Power Utilization (Offline)}
\label{tbl:benchmark-exp1}
\begin{tabular}{cc|ccc|c|c|c}
 \hline 
 \hline
 \multicolumn{2}{c|}{Hardware} & \multicolumn{3}{c|}{Power ($mW$)} & \multirow{2}{*}{Latency per Timestep ($\mu s$)} & \multirow{2}{*}{Energy per Timestep ($\mu J$)} & \multirow{2}{*}{Energy-Delay Product (nJs)} \\
  { } & & Static & Dynamic & Total &  &  &  \\
 \hline 
\multirow{3}{*}{Loihi} & x86 cores & 0.19 & 21.1 & 21.3 & - & 0.14 & - \\
& neuron cores & 11.7 & 0.64 & 12.4 & - & 0.08 & - \\
& total & 11.9 & 21.7 & 33.7 & 231 & 0.22 & 51.8 \\
\hline
GPU & total & 3594 & 56296 & 59890 & 21 & 399 & 8431 \\
 \hline 
 \hline 
\end{tabular}
\end{table*}

\begin{table*}
\centering
\caption{Inference Speed and Power Utilization (Real-world Simulation)}
\label{tbl:benchmark-exp2}
\begin{tabular}{cc|ccc|c|c|c}
 \hline 
 \hline
 \multicolumn{2}{c|}{Hardware} & \multicolumn{3}{c|}{Power ($mW$)} & \multirow{2}{*}{Latency per Timestep ($\mu s$)} & \multirow{2}{*}{Energy per Timestep ($\mu J$)} & \multirow{2}{*}{Energy-Delay Product (nJs)} \\
  { } & & Static & Dynamic & Total &  &  &  \\
 \hline 
\multirow{3}{*}{Loihi} & x86 cores & 0.19 & 20.1 & 20.3 & - & 0.14 & - \\
& neuron cores & 11.9 & 0.15 & 12.1 & - & 0.08 & - \\
& total & 12.1 & 20.23 & 32.3 & 1039.9 & 0.22 & 224 \\
\hline
GPU & total & 3594 & 58336 & 61930 & 1045.6 & 412.9 & 431708 \\
 \hline 
 \hline 
\end{tabular}
\end{table*}

We then ran a second experiment that simulates accurately the real-world setting. The setup is identical to the first experiment, with two changes:

\begin{enumerate}
    \item The x86 cores are artificially slowed down to match the 1ms timestep duration of the data.
    \item An artificial delay of 0.15s is introduced to the dataset fetch for the GPU, to simulate waiting for the full window of data before it is able to perform inference.
\end{enumerate}

The results are presented in Table~\ref{tbl:benchmark-exp2}. We observe that the inference speeds on the GPU and Loihi are comparable in this scenario, but the Loihi consumes 1927x less power than the GPU. The Loihi's slightly faster inference speed is likely due to the chip being able to process the spikes as they arrive, while the GPU needs to wait for the full window of spiking data to arrive before performing inference. However, we note that the GPU remains more efficient for offline, batched processing of spiking data, and the latency difference would be further compounded by passing data in larger batch sizes to the GPU.

\end{document}